\documentclass[a4paper, 10pt, conference]{ieeeconf}      
\usepackage{FG2021}
\usepackage{mathtools,amssymb}
\usepackage{footnote}
\usepackage{diagbox}
\usepackage{caption}
\usepackage{subcaption}
\usepackage{multirow}
\usepackage{tabularx}
\usepackage{amsfonts}
\usepackage{booktabs}
\usepackage{siunitx}

\FGfinalcopy 

\IEEEoverridecommandlockouts                              
\def\FGPaperID{****} 

\title{\LARGE \bf
Your ``Attention" Deserves Attention: A Self-Diversified Multi-Channel Attention for Facial Action Analysis
}

\author{\parbox{16cm}{\centering
    {\large Xiaotian Li$^1$, Zhihua Li$^1$, Huiyuan Yang$^3$, Geran Zhao$^2$ and Lijun Yin$^1$}\\
    {\normalsize
    $^1$ Department of Computer Science, State University of New York at Binghamton, New York, USA\\
    $^2$ Department of Mathematical Science, State University of New York at Binghamton, New York, USA\\
    $^3$ Department of Electrical and Computer Engineering, Rice University, Texas, USA}}
}

\begin{document}

%
%
%




\IEEEoverridecommandlockouts\pubid{\makebox[\columnwidth]{978-1-6654-3176-7/21/\$31.00~\copyright{}2021 IEEE \hfill}
\hspace{\columnsep}\makebox[\columnwidth]{ }}

\ifFGfinal
\thispagestyle{empty}
\pagestyle{empty}
\else
\author{Anonymous FG2021 submission\\ Paper ID \FGPaperID \\}
\pagestyle{plain}
\fi
\maketitle

\begin{abstract}

Visual attention has been extensively studied for learning fine-grained features in both facial expression recognition (FER) and Action Unit (AU) detection. A broad range of previous research has explored how to use attention modules to localize detailed facial parts (e,g. facial action units), learn discriminative features, and learn inter-class correlation. However, few related works pay attention to the robustness of the attention module itself. Through experiments, we found neural attention maps initialized with different feature maps yield diverse representations when learning to attend the identical Region of Interest (ROI). In other words, similar to general feature learning, the representational quality of attention maps also greatly affects the performance of a model, which means unconstrained attention learning has lots of randomnesses. This uncertainty lets conventional attention learning fall into sub-optimal. In this paper, we propose a compact model to enhance the representational and focusing power of neural attention maps and learn the ``inter-attention" correlation for refined attention maps, which we term the ``Self-Diversified Multi-Channel Attention Network (SMA-Net)". The proposed method is evaluated on two benchmark databases (BP4D and DISFA) for AU detection and four databases (CK+, MMI, BU-3DFE, and BP4D+) for facial expression recognition. It achieves superior performance compared to the state-of-the-art methods.

\end{abstract}

\section{INTRODUCTION}

Both automatic facial expression recognition (FER) and facial action unit (AU) detection play a vital role in revealing human affective states. Over the past few decades, deep learning has achieved remarkable performance in automatic facial action analysis as its ability to capture high-level abstractions through hierarchical architectures of multiple non-linear transformations and representations \cite{Relatedwork_1}.

Attention is a behavioral and cognitive process of focusing selectively on a discrete aspect of information, while ignoring other perceptible information, playing an essential role in human cognition. The Neural Attention Network (NAN) was originally used for machine translation in natural language processing (NLP). After that, the concept was extended to the field of computer vision \cite{Introduction_9}. Current facial works \cite{Introduction_5,Introduction_6,Introduction_7,Experiments_11,Relatedwork_15,Introduction_14}, disentangle a global face into smaller parts (e,g. mouth of a face) for fine-grained classification. Specifically, they adopt multiple spatial attention to localize the regions of interest (ROIs) and learn the discriminative facial features. Although some progress has been made, this design has paid a price on the other hand. First of all, the strong dependencies and mutual exclusive relation among local parts are ignored. For instance, the AU1 and AU2 are usually co-existence due to the constraints of facial muscles. Thus, decoupling a global face into isolated parts undermines the relationship between AUs. Second, lacking a global view can weaken the learning performance, and significantly impair the representational power of the network. More importantly, paying attention to specific facial parts might capture irrelevant information. For instance, both AU9 and AU10 have the same muscle activation on levator labii superioris which is visually indivisible. After a careful investigation of related works \cite{Introduction_7, Experiments_15, Experiments_9, Relatedwork_15}, we observed that the displayed attention maps between (AU4 and AU7), (AU12, AU14, and AU15), and (AU23 and AU24) have obvious regional overlap. Consequently, the ambiguous localization of facial actions leads the discriminative features less efficient. Thus, we argue that designing a more robust attention framework is still a challenging task for fine-grained facial analysis.

\begin{figure}
\begin{center}
\includegraphics[width=0.93\linewidth]{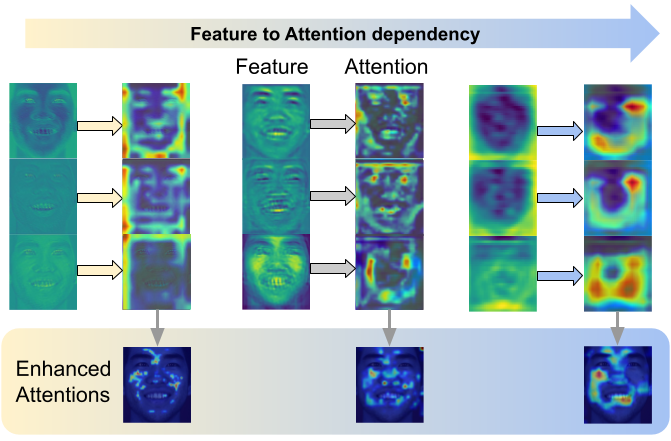}
\caption{Overview of the ``Feature-to-Attention" dependency. Three sets of multi-channel attention maps in different stages of the deep model from shallow to deep. In each set, the generated attention maps are highly dependent on the input feature maps and present distinct attention performance. Our model takes advantage of the ``Feature-to-Attention" dependency to construct enhanced attention with diverse representations for automatic facial analysis.}
\label{pipeline_module}
\end{center}
\end{figure}

Multi-head attention \cite{Introduction_17} is a mechanism that can boost the performance of the vanilla self-attention layer in NLP. In order to address the transformer's limitation on computer vision, ViT \cite{Methodology_5} splits an image into fixed-sized patches to linearly embed them with position embedding. These works have demonstrated that the multi-head design is beneficial for learning enhanced attention maps. Inspired by them, we disentangle a collection of feature maps extracted by a convolution neural network (CNN) and map them to several sub-spaces for learning the corresponding spatial attention. Through experiments, we found the spatial attention with diverse initialization can yield different performances when seeking the identical Region of Interest (ROI). Fig.\ref{pipeline_module} depicts a clear dependency between the feature maps and attention maps. In other words, similar to general feature learning, the representational quality also greatly affects the performance of attention learning. Fig.\ref{attention_map_2} depicts the visual comparison between conventional fine-grained attention learning and our method.

In this paper, we propose a compact model named ``Self-Diversified Multi-Channel Attention Network (SMA-Net)" to enhance the representational power of attention maps. SMA-Net consists of three key components: (1) ``Feature-to-Attention channel" (F2A) are jointly initialized and trained with different deep features mapped from the local receptive field, for enhancing the quality of spatial attention encoding. In each channel, we deploy spatial attention to learn the facial ROIs and learn the pixel-level relations for facial actions in a holistic way. The performance of the attention map generated under different channels also shows differences. (2) multiple losses encourage the attention channels to focus on identical locations yet learn diverse representations of attention maps. (3) ``Attention-Above-Attentions" (AAAtention), similar to typical channel-wise attention, is applied to selectively emphasize optimal attention channels while suppressing less useful ones, and model the inter-relation among F2A channels, which we term the ``inter-attention" correlation. Our contribution lies in three-fold:
\begin{itemize}
\item We found the robustness issue of attention-based learning which is overlooked in conventional facial action analysis. We provide a new understanding of the correlation between CNN-based feature maps and attention maps, yielding a novel multi-channel attention framework.
\item The proposed network can explicitly strengthen the representational and focusing ability of CNN-based attention without introducing many trainable parameters or regularizers.
\item This design can be applied to both multi-label and multi-class recognition tasks. The result shows it outperforms the peer state-of-the-art algorithms on two datasets (BP4D and DISFA) for facial action unit detection and achieves competitive performance than other benchmark approaches on four databases (CK+, MMI, BU3DFE, and BP4D+) for facial expression recognition. Additional experiments show that it achieves obvious improvement compared with conventional spatial attention, channel-wise attention, and mixture attention.
\end{itemize}

\section{RELATED WORK}

\begin{figure*}[ht]
\begin{center}
\includegraphics[width=0.93\linewidth]{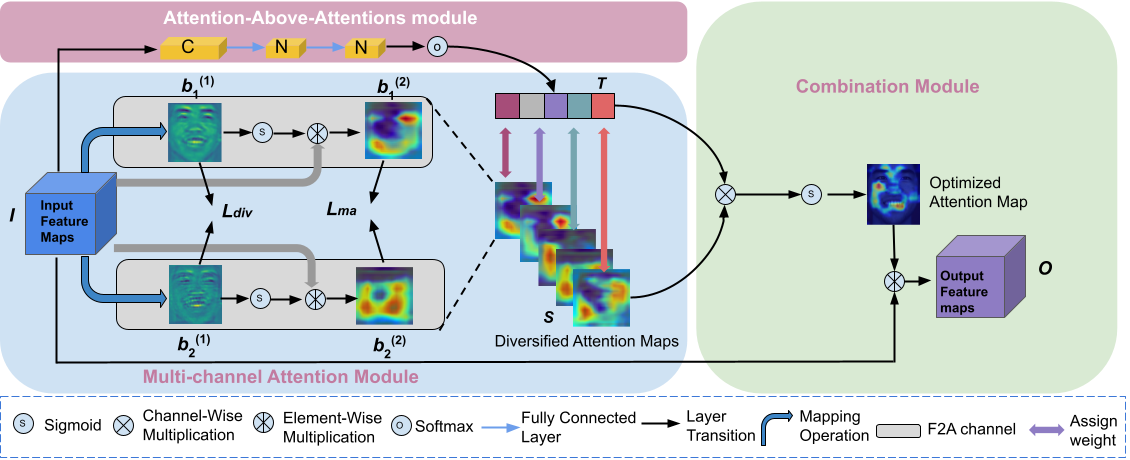}
\caption{Overview of the key components in our Self-Diversified Multi-Channel Attention Network (SMA-Net). The model takes the extracted feature block as input. After mapping the feature block into $N$ attention channels (here $N$ = 2), the diversity losses $l_{div}$ and multi-attention losses $l_{ma}$ jointly encourage the channels to learn the same active parts yet present diverse representational patterns. The diversified attention maps are evaluated and aggregated into the next stage for getting the optimized attention map. Noting that ``Self-Diversified" means the representational diversity of attention maps is generated with the CNN-based channel-mapping mechanism and an unsupervised loss function $l_{div}$ without using any auxiliary information.}
\label{Overview_module}
\end{center}
\end{figure*}

\subsection{Facial action analysis}
Facial expression and action unit features can be roughly categorized into two classes: shallow features and high-level features. In this article, we mainly focus on deep-features-based models. Mollahosseini et al. \cite{Relatedwork_2} presented a new deep neural network (DNN) architecture to deal with the FER task. \cite{Relatedwork_3, Relatedwork_4} demonstrate the deep features perform an impressive function on both AU multi-label classification and intensity estimation. Predefined attention was a popular method to learn the ROIs regarding both AU detection and FER. \cite{Relatedwork_11} proposed the EAC-Net by using predefined attention, whose location is derived by face landmarks, to enhance and crop the ROIs of AUs. Sanchez et al. \cite{Relatedwork_13} adopted the Gaussian distribution to predefine the central location of specific AU according to the fixed landmark for AU intensity estimation. However, due to lacking the accurate definition of the contour and location bias cross identity, ethnicity, gender, age, and facial expressions, the predefined ROIs fall into sub-optimal and lose the adaptation during the recognition process. To tackle this issue, \cite{Introduction_7, Relatedwork_7, Relatedwork_15, Introduction_6, Introduction_5, Experiments_11} apply the self-attention mechanism to discover the meaningful facial locations with more adaptations for FER, AU detection, AU intensity estimation, and face alignment tasks. In addition, self-attention is also widely used for capturing the correlations among AUs \cite{Introduction_7}, spatio-temporal relation \cite{Relatedwork_18}, and modality-level relation \cite{Relatedwork_19}.

\subsection{Attention mechanism}
Attention emerges in CNN to filter information and allocate weights to the neural network efficiently. Attention mechanisms direct the operational focus of deep neural networks to areas where there is more saliency information. Attention mechanism has been successfully applied to deep CNN-based image enhancement methods. Zhang et al. \cite{Relatedwork_20} proposed a residual channel attention network (RCAN) in which residual channel attention blocks (RCAB) allow the network to focus on the more informative channels. 
Huet al. \cite{Relatedwork_21} proposed a Squeeze-and Excitation (SE) block to improve the quality of representations produced by a network by explicitly modeling the inter-dependencies between the channels of its convolutional features. Woo et al. \cite{Experiments_28} proposed channel attention (CA) and spatial attention (SA) modules to exploit both inter-channel and inter-spatial relationships of feature maps. ViT \cite{Introduction_18} splits an image into fixed-sized patches to linearly embed them with position embedding. Our method differs from all related works significantly in the following facts: (1) existing works focus on refining the deep feature, while our method does not. Instead, our method is proposed to enhance the attention itself. (2) In contrast to \cite{Introduction_7} that uses attention for learning the inter-AU correlation, our method utilizes the attention mechanism at two different levels: Inter-AU (pixel level) and Inter-attention (channel level). Inter-AU correlation is nested within Inter-attention correlation, thus capturing the high-dimensional information for facial analysis. (3) Our method does not need to disentangle a face for focusing on each facial part independently, which greatly reduces the designing complexity. (4) The proposed design can be extended for discrete facial expression recognition, which is more flexible than general AU-based works.

\section{Methodology}

In this section, we elaborate on the structure and key components of SMA-Net. The framework of our model is demonstrated in Fig.\ref{Overview_module}.

\subsection{Multi-channel attention module}

The feature map block $I$ is initially extracted by ResNet18 as the input of this module. Fig.\ref{Overview_module} shows the overview of the proposed architecture. Inspired by the multi-head attention design \cite{Methodology_5, Introduction_17}, this module splits into several sub-channels or sub-layers which accommodate the deep feature maps under the different channels of a block. For the initialization of each channel, our method does not follow the previous principles such as using global pooling or average pooling. We expect the multiple attention maps can show diversity from the initial stage. In addition, considering the common symptoms of over parameterization for conventional multi-head attention works \cite{Methodology_7}, this module reduces the channel dimension from $C$ to $N$ by applying a simple convolutional operation, where $N$ is the number of attention channels. After channel mapping operation, we get the feature maps $b_{n}^{1}$, where $n$ denotes the n-th (from 1 to $N$) channel and $1$ means the first stage of the F2A channel. The channel number $N$ can be set as a hyper-parameter of the model. In Sec. \ref{N-number}, we have investigated and discussed how the channel number $N$ affects the recognition performance.

Zhu et al. \cite{Introduction_7} adopt spatial attention to capture AU-related local features and the pixel-level relation within independent AUs. Different from his work, our method does not disentangle a face to multiple facial parts. We treat the face as a holistic target to prevent ambiguity caused by any invalid segmentation. Besides, the spatial attention in each F2A channel can effectively learn the AU-related feature and the pixel-level inter-AU correlation. To be specific, we further extract $b_{n}^{1}$ by using another convolutional layer $F_{n}^{1}$ to get $b_{n}^{1}$. The spatial attention map $S_{n}$ in each branch is computed as:
\begin{equation}
S_{n}=I\circledast\sigma(F_{n}^{1}(I))
\end{equation}
where $\sigma$ denotes the sigmoid function and $\circledast$ denotes the element-wise multiplication, $I\in \mathbb{R}^{C \times  H \times  W}$, and $S_{n}(I)\in \mathbb{R}^{C \times  H \times  W}$. The convolutional operation $F_{n}^{1}$ takes 7 $\times$ 7 as the filter size, and the input/output channel is 1/1. Noting that the sigmoid function is used for attention activation in our F2A channels.

To enable our model to discover diverse attention patterns towards the same object in F2A channels, we design the diversity loss $l_{div}$ and multi-attention loss $l_{ma}$  respectively. The design is inspired by \cite{Methodology_7}. First, we expect the mapped features used for initialization to have more discriminative information (not task-dependent) and share less overlapping in image representation. To fulfill this goal, we apply a geometric constrain to encourage the multiple attention to learn divergent initialization distribution in an unsupervised way. The $l_{div}$ is formulated by:
\begin{equation}
l_{div}=\sum_{n\in N}^{}m^{n}max\left \{ 0, \widehat{m}-\delta \right \}
\end{equation}
where $\widehat{m}^{n}=maxm_{k}^{n}$ denotes the maximum of the other attention maps at channel $n$ and $\delta $ means a margin. It can be considered as the inner product of two flattened matrices, which measures the distance of two feature maps. The multi-attention losses are basically the same as the loss function of classification depending on the task (FER or AU detection). The minor difference is $l_{ma}$ exists in every attention channel and the corresponding bypass layers are used for the loss calculation. Ideally, our model aims to use $l_{ma}$ to activate the same facial parts in multiple F2A channels.

\subsection{Attention-Above-Attentions module}

The Attention-Above-Attentions (AAAtention) module, in Fig.\ref{Overview_module}, is constructed as a global attention system to automatically weigh the performance of F2A channels. Domain Attention in \cite{Relatedwork_6} use the squeeze-and-excitation module \cite{Methodology_1} to produce domain-sensitive weighting module for universal object detection. Different from \cite{Relatedwork_6}, our module aims to selectively emphasize optimal attention channels while suppressing less useful ones, and capture the inter-relation among F2A channels. $T$ in Fig. \ref{Overview_module} stands the correlation map of F2A channels. This module applies average-pooling to remove the spatial information of $I$. The average-pooling operation is denoted as a descriptor $F_{avg}$. Then the channel-wise attention weight $T\in \mathbb{R}^{1 \times  1 \times  N}$ can be computed as:
\begin{equation}
T=\varphi(W^{1}(W^{2}(F_{avg}(I))))
\end{equation}
Where $\varphi$ denotes the softmax function. $W^{1}\in N$ and $W^{2}\in N$ denote the weight of the full-connected layer as a MLP framework. Note that no ReLU activation is included in the module.

\begin{figure}[ht]
\begin{center}
\includegraphics[width=0.95\linewidth]{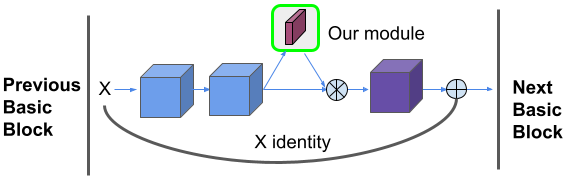}
\caption{The location of the multi-attention module in the basic-block of ResNet18. It is located before the summation with the identity branch in a basic block of ResNet.}
\label{location}
\end{center}
\end{figure}

\subsection{Combination module}

The combination module makes element-wise multiplication with the channel-wise attention map $T$ and the aggregated multi-channel attention maps $S$. If we let $\ast$ denote the channel-wise multiplication operation, the optimized attention map $A\in \mathbb{R}^{1\times H\times W}$ can be computed as:
\begin{equation}
A=\sigma (F_{avg}(T*S))
\end{equation}
The refined output feature $O\in \mathbb{R}^{C\times H\times W}$ can be expressed as:
\begin{equation}
O=A\circledast I
\end{equation}
where $\circledast$ denotes the element-wise multiplication.

\subsection{Backbone network and overall objective}
We choose ResNet-18 as the backbone network. It's worth noting that the parameter scale of the proposed model is slightly increased by only $\sim$4\% from the original network. Compared with the most popular architectures based on ResNet-50, Inception-v4 \cite{Relatedwork_9},  ResNeXt-101 \cite{Methodology_3}, EfficientNet-B4 \cite{Methodology_4}, ViT-H/14 \cite{Methodology_5} and NFNet-F0 \cite{Methodology_6} towards image classification, the proposed framework can save around 50\% to 90\% cost on the parameter scale. It is an end-to-end trainable network and no prior knowledge is used. Fig.\ref{location} shows the exact location of our key module. The modules are placed in every basic-block of ResNet18 to make sure its consistent and extensive inference towards AU detection. They are placed in the first two basic-block in the backbone network when applying to the FER task.

Considering the data imbalance issues from skewing the training process and affect the performance of the model. We choose weighted BCE logits loss and the function can be described as:
\begin{equation}
l(x,y)=L=\left \{ l_{1},...,l_{N} \right \}^{T}
\end{equation}
\begin{equation}
l_{n}=-w_{n} [y_{n} log\sigma \left ( x_{n} \right ) +\left ( 1-y_{n} \right )log\left ( 1-\sigma \left (x_{n}\right ) \right ) ]
\end{equation} 
where $N$ is the batch size. and $l(x,y)=mean(L)$.

For the facial expression recognition task, the cross-entropy loss is selected as the loss function loss which can be described as:
\begin{equation}
l(x,class)=-log\left \{\frac{ exp(x[class])}{\sum _{j}exp(x[j])} \right \}
\end{equation}

The overall function can be can be described as:
\begin{equation}
l_{all} = l_{cla} + \alpha l_{div} + \lambda l_{ma}
\end{equation}
where $\alpha$ and $\lambda$ are the balance factors, and $l_{cla}$ denotes the loss function for classification depending on the specific task (FER or AU detection).

\section{Experiments}
\subsection{Datasets and Settings}
The proposed framework is evaluated on two benchmark datasets: BP4D \cite{Experiments_1} and DISFA \cite{Experiments_2} for action unit detection. In addition, it is evaluated on four FER databases: Extended Cohn-Kanade (CK+) \cite{Experiments_4}, MMI \cite{Experiments_5}, BU-3DFE \cite{Experiments_6}, and BP4D+ \cite{Experiments_7}.

\textbf{BP4D} and \textbf{DISFA}: BP4D contains 41 subjects and around 140,000 frames captured under laboratory environments. There are 27 subjects and around 130, 000 frames in the DISFA database. Following the previous settings of DISFA, the frames with intensities equal to or larger than 2 are regarded as AU occurrences while the rest are absent. We divide both datasets into subject-independent 3 folds and report the performance through cross-validation for a fair comparison with other algorithms. We observed a severe data imbalance issue in DISFA which affects the evaluation significantly for the less representative AUs. To tackle such an issue, we apply a selective data re-sampling technique that enhances the less occurred AUs. The selective oversampling process duplicates the minority classes in the training set. Meanwhile, setting the threshold $p$ can control unnecessary oversampling from other majority classes which may cause overfitting.

\textbf{CK+}, \textbf{MMI}, \textbf{BU-3DFE} and \textbf{BP4D+}: CK+ contains 593 video sequences from 123 subjects. We use the last three frames of each sequence with the provided label, which constitutes a set with 981 images. MMI consists of 236 image sequences from 31 subjects. Each sequence is labeled as one of the six prototypical facial expressions. We selected three frames in the middle of each sequence as peak frames in frontal view. This builds a dataset with 624 images. BU-3DFE contains 2,500 pairs of static 3D face models and texture images from 100 subjects with a variety of ages and races. During the experiment, only the 2D texture images with high-intensity expressions are used. BP4D+ is a spontaneous emotion corpus with 140 subjects. Following the work \cite{Experiments_19}, we select 2468 frames from 72 subjects on four tasks based on the FACS codes. The 2D texture images are used and four kinds of expressions (i.e., happiness, surprise, pain, and neutral) are regarded as the ground truth. Among the four datasets, the images are split into 10 folds and the subjects in any two subsets are mutually exclusive.

\subsection{Implementation details}
\label{N-number}
\begin{figure}
\begin{center}
\includegraphics[width=0.95\linewidth]{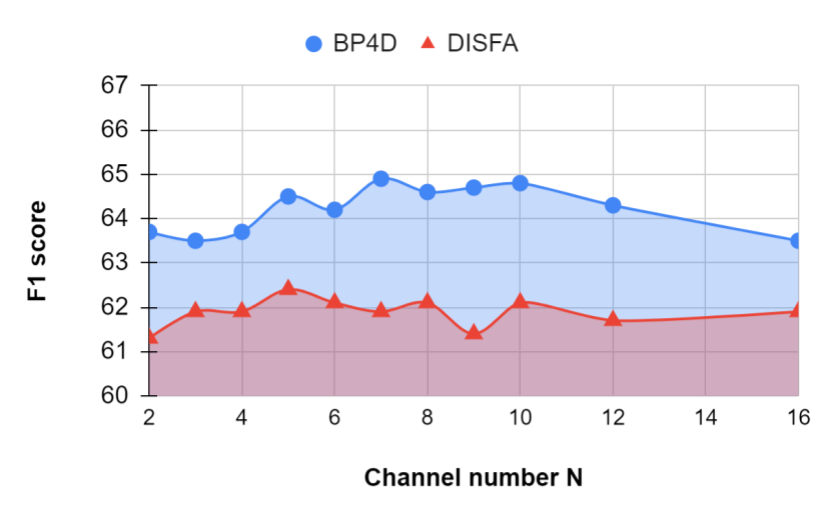}
\caption{The relation between the $F_{1}$ score and the channel number $N$ on BP4D and DISFA.}
\label{branch_number}
\end{center}
\end{figure}

For the AU detection, we first process the image by the cropping operation to cut off redundant area which is not relevant to the face recognition. Then the images are resized to 256*256*3 (H*W*C) to fit the model. Each of the training images is randomly rotated (-45 to 45 degrees), flipped horizontally (50\% possibility), and with color jitters (saturation, contrast, and brightness). The backbone network is pre-trained on ImageNet \cite{Experiments_3}. We choose SGD as the optimizer with an initial learning rate of 0.01. It is decreased to 0.001 after the first 2 epoch training. The weight decay and momentum are set as 0.0001 and 0.9 respectively. The number $N$ of attention branches is set as 7 and 5 on BP4D and DISFA respectively. About how the channel number $N$ affects the recognition result, we observe the performance curve with different number $N$ in Fig.\ref{branch_number}. It shows the result improves roughly as the number increases until it reaches a certain threshold. We infer that a high number $N$ will not only cause the saturation of branch diversity but also generate overlapping attention maps that may dominate the model training and drive the learning process to omit valuable information from other branches.

For the FER, we cropped the image and resized it to 64*64*3 (H*W*C). Before training, the images are randomly horizontal flipped (50\% possibility), rotated (-15 to 15 degrees), and with color jitters (saturation, contrast, hue, and brightness). We choose SGD as the optimizer with an initial learning rate of 0.01. The learning decay rate is 0.99 and it decays every 10 epoch until the 100 epoch is finished. The number $N$ of attention branches is set as 7.

\subsection{Results and Discussions}
\textbf{Action unit detection}: we compared the proposed network with other state-of-the-art algorithms. In Table \ref{tab:f1_BP4D} and Table \ref{tab:f1_DISFA}, the $F_{1}$ score matrices, which can be described as $F_{1}=2\cdot \frac{Precision\cdot Recall}{Precision+Recal}$, is applied to estimate the performance in AU detection. Our method outperforms other benchmark algorithms, including the attention-based models, on both BP4D and DISFA. It shows our model achieves the superior performance for AU detection, especially we achieve the best $F_{1}$ in eight individual AUs(AU1, 6, 7, 10, 14, 15, 17, and 24) on BP4D.

\begin{figure*}[ht]
\begin{center}
\includegraphics[width=1\linewidth]{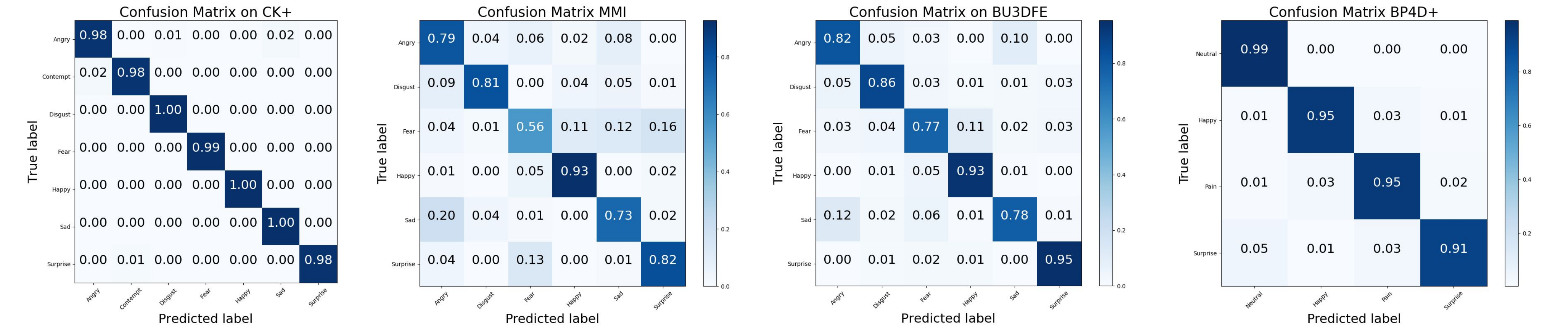}
\caption{The summary of the confusion matrix on CK+, MMI, BU3DFE and BP4D+.}
\label{confusion_matrix}
\end{center}
\end{figure*}

\textbf{Facial expression recognition}: we extend the evaluation to the facial expression recognition task. In Table \ref{tab:acc_CK_MMI} and Table \ref{tab:acc_BU3DFE_BP4D}, SMA-Net achieves competitive performance in terms of the accuracy against the state-of-the-art models on four datasets. The confusion matrix in Fig.\ref{confusion_matrix} based on four datasets shows that fear, angry and sad are relatively more difficult to recognize due to the high socialized and complex attributes of human behavior.

\begin{table}\renewcommand\arraystretch{1}
\caption{ F1 score of the state-of-the-art models on BP4D. Bold numbers indicate the best performance.}
\begin{center}
\resizebox{0.50\textwidth}{23mm}{
 \begin{tabular}{|c|cccccc|c|}

\hline
AU  &   DSIN \cite{Experiments_8} & JAA \cite{Experiments_9} & ARL \cite{Introduction_7}   & LP \cite{Experiments_11} & SRERL \cite{Introduction_12} & RoIGC \cite{Experiments_12} & Ours \\\hline  
1   &     51.7   &  47.2   &  54.8   &  43.4 & 46.9 & 52.6 & \textbf{56.5}  \\ \hline   
2   &      40.4   &  44.0   &  39.8   &  38.0 & 45.3 & \textbf{47.0} & 45.1  \\ \hline
4   &     56.0   &  54.9   &  55.1  & 54.2 & 55.6 & \textbf{61.4} & 57.0\\ \hline
6 &      76.1   &  77.5   &  75.7   &  77.1 & 77.1 & 76.8 & \textbf{79.5} \\ \hline
7  &    73.5   & 74.6   & 77.2   &  76.7 & 78.4 & 79.2 & \textbf{79.5} \\ \hline
10   &     79.9  & 84.0   & 82.3  &  83.8 & 83.5 & 83.5 & \textbf{84.5} \\ \hline
12  &       85.4   & 86.9   & \textbf{86.6}   & 87.2 & 87.6 & 88.6 & 86.4 \\ \hline
14  &     62.7   & 61.9   & 58.8   & 63.3 & 63.9 & 60.4 & \textbf{66.1}  \\ \hline
15  &       37.3   & 43.6   & 47.6   & 45.3 & 52.2 & 49.3 & \textbf{55.8}  \\ \hline
17  &      62.9   & 60.3   & 62.1   & 60.5 & 63.9 & 62.6 & \textbf{64.2}  \\ \hline
23  &       38.8   & 42.7   & 47.4   & 48.1 & 47.1 & \textbf{50.8} & 48.7 \\ \hline
24  &       41.6   & 41.9   & 55.4   & 54.2 & 52.3 & 49.6 & \textbf{56.8}  \\ \hline
Avg. &       58.9   & 60   & 61.1   & 61.0 &  62.9 & 63.5 & \textbf{64.9}  \\ \hline

\end{tabular}}
\label{tab:f1_BP4D}
\end{center}
\end{table}

\begin{table}\renewcommand\arraystretch{1}
\caption{ F1 score of the state-of-the-art models on DISFA. Bold numbers indicate the best performance.}
\begin{center}
\resizebox{0.49\textwidth}{19mm}{
 \begin{tabular}{|c|cccccc|c|}

\hline
AU  &    DSIN \cite{Experiments_8} & JAA \cite{Experiments_9} & ARL \cite{Introduction_7}  & LP \cite{Experiments_11} & SRERL \cite{Introduction_12} & RoIGC \cite{Experiments_12} & Ours \\\hline  
1   &      42.4   &  43.7   &  43.7   &  29.9 & 45.7 & \textbf{55.0} & 53.4  \\ \hline   
2   &       39.0   &  46.2   &  42.1   &  24.7 & 47.8 & \textbf{63.0} & 54.2  \\ \hline
4   &     68.4   &  56.0   &  63.6  & 54.2 & 72.7 & \textbf{74.6} & 64.0\\ \hline
6  &   28.6   &  41.4   &  41.8   &  46.8 & 47.1 & 45.3 & \textbf{57.0} \\ \hline
9  &      46.8   & 44.7   & 40.0   &  \textbf{76.7} & 49.6 & 35.2 & 47.0 \\ \hline
12  &    70.8   & 69.6   & 76.2   & \textbf{87.2} & 72.9 & 75.3 & 76.6 \\ \hline
25  &   90.4   & 88.3   & 95.2   & 63.3 & \textbf{93.8} & 93.5 &  92.0 \\ \hline
26  &     42.2   & 58.4   & \textbf{66.8}   & 45.3 & 65.0 & 54.4 &  55.2  \\ \hline
Avg. &   53.6   & 56.0   & 58.7   & 61.0 &  56.9 & 62.0 &  \textbf{62.4}\\ \hline

\end{tabular}}
\label{tab:f1_DISFA}
\end{center}
\end{table}

\begin{table}\renewcommand\arraystretch{1}
\caption{ Accuracy of the state-of-the-art models on CK+ and MMI. Bold numbers indicate the best performance. Numbers with underline means sub-optimal performance.}
\begin{center}
\resizebox{0.49\textwidth}{20mm}{
 \begin{tabular}{|c|c|c||c|c|c|}

\hline
\textbf{CK+}  &  Data   & Accuracy &  \textbf{MMI}  &  Data  & Accuracy  \\\hline  
HOG 3D \cite{Experiments_14}  &   sequence &  91.44 & HOG 3D \cite{Experiments_14}  &   sequence & 60.89 \\ \hline
STM-Explet \cite{Experiments_16} & sequence   &  94.19 & STM-Explet \cite{Experiments_16} & sequence   & 75.12\\ \hline
IACNN \cite{Experiments_17} &  image  &  95.37 & IACNN \cite{Experiments_17} &  image  & 70.24 \\ \hline
DTAGN \cite{Experiments_18} & sequence   &  97.25 & DTAGN-Joint \cite{Experiments_18} & sequence   & 71.55\\ \hline
MSFLBP \cite{Experiments_26} & image   &  \underline{99.12} & - & -   & - \\ \hline
FMPN \cite{Experiments_27} & image   &  98.06 & FMPN \cite{Experiments_27} & image   & \underline{82.74}\\ \hline
Attention CNN \cite{Experiments_25} & image   &  98.68 & - & -   & - \\ \hline
DeRL \cite{Experiments_19} & image   &  97.30  & DeRL \cite{Experiments_19} & image   & 73.23  \\ \hline
ResNet18 & image   &  97.67  & ResNet18 & image   & 75.62 \\ \hline
\textbf{Our SMA-Net}  &  image   &  \textbf{99.17}  & \textbf{Our SMA-Net}  &  image   & \textbf{82.75}\\ \hline

\end{tabular}}
\label{tab:acc_CK_MMI}
\end{center}
\end{table}

\begin{table}\renewcommand\arraystretch{1}
\caption{ Accuracy of the state-of-the-art models on BU3DFE and BP4D+. Bold numbers indicate the best performance. Numbers with underline means sub-optimal performance.}
\begin{center}
\resizebox{0.49\textwidth}{20mm}{
 \begin{tabular}{|c|c|c||c|c|c|}
\hline
\textbf{BU3DFE}  &  Data   & Accuracy & \textbf{BP4D+}  &  Data   & Accuracy \\\hline 
Berretti et al.\cite{Experiments_21}  &   3D &  77.54& - & -   & - \\ \hline
Yang et al.\cite{Experiments_22} & 3D   &  84.80 & - & -   & -\\ \hline
Lo et al.\cite{Experiments_23} &  2D + 3D  &   \textbf{86.32} & - & -   & -\\ \hline
Lopes \cite{Experiments_24} & 2D   &  72.89 & - & -   & -\\ \hline
FERAtt \cite{Experiments_19} & 2D   &  85.15 & - & -   & - \\ \hline
DeRL \cite{Relatedwork_14} & 2D   &  84.17 & DeRL \cite{Experiments_19} & 2D   &  81.39 \\ \hline
- & -   &  -  & SEnet18 \cite{Methodology_1}   &   2D image &  94.25\\ \hline
- & -   &  -  & CBAM \cite{Experiments_28}   &   2D image &  94.20\\ \hline
ResNet18 & 2D    &  84.16  & ResNet18   &   2D image &  93.74\\ \hline
\textbf{Ours SMA-Net}  &  2D    &  \underline{85.42} & \textbf{Ours SMA-Net}  &  2D   &   \textbf{95.41}\\ \hline
\end{tabular}}
\label{tab:acc_BU3DFE_BP4D}
\end{center}
\end{table}

\subsection{Ablation study}
In this section, we investigate the effectiveness of each module in the proposed model. Table \ref{tab:ablation_study} shows the ablation study in terms of the $F_{1}$ score on BP4D and DISFA. The baseline model is ResNet-18.

\textbf{Effect of Key modules:} The F2A has a collection of channels to construct the dependency between feature maps to attention maps. By inheriting the information-rich clues from the feature maps, F2A significantly affects the attention learning. From Table \ref{tab:ablation_study}, we can observe the average $F_{1}$ score increases from 60.8 to 62.7 by adding F2A module to baseline. ``multi-channel" in Table \ref{tab:ablation_study} means applying multiple spatial attention (average-pooling based initialization) without diversifying the pattern of attention branches. It proves, without diversifying the attention initialization, the performance of multi-head design has been weakened to a certain extent. Attention-Above-attentions (AAA) module shows limited ability to refine the model. We argue that the AAA, as typical channel-wise attention, misses lots of spatial information which is more sensitive to facial action analysis. However, by combining the F2A and AAA modules, our model achieves obvious performance improvement from the baseline by increasing 5.2 $F_{1}$ score (3.6 on BP4D and 6.7 on DISFA). For a fair comparison, We also applied the design that splitting 12 AU-related attention (F2A+AAA+local-AU) to each corresponding AU respectively for a finer local region learning, inspired by \cite{Relatedwork_15}. We split 12 attention branches from deep features $I$. Each independent attention branch is employed to refine the local features and predict a single AU occurrence, in which the multi-label classification issue is constructed as a 12 binary classification problem. The result shows that the AU-related local attention design can not surpass our model under the same circumstance. We believe it is due to the lacking a robust attention learning mechanism. Specifically, although the conventional multi-head structure allows the model to pay attention to different parts, it can not guarantee whether each branch has strong executive power. In contrast, our model can avoid the issue by querying AU-specified attention from multiple channels.

\textbf{Effect of loss functions:} The bottom part of Table \ref{tab:ablation_study} shows how the proposed loss functions affect our model. Compared with the baseline (F2A+AAA) the $l_{ma}$ loss improves slightly as it can apply the activation functions deeply in every attention channel. The result inferred from the $l_{div}$ is better, and we get the best performance when combining the inferences of $l_{ma}$ and $l_{div}$, which improve the baseline by an average of 1.3 $F_{1}$ score.

\begin{table}\renewcommand\arraystretch{1}
\caption{ Ablation study on BP4D and DISFA. Bold numbers indicate the best performance.}
\begin{center}
\resizebox{0.36\textwidth}{23mm}{
\begin{tabular}{SSSS} \toprule
    {Method} & {BP4D} & {DISFA} & Avg \\ \midrule
    {Baseline}                  & 60.8 & 53.6 &  57.2 \\ \midrule
    {multi-channel}             & 62.3   & 56.5 &  59.4    \\
    {F2A}                       & 62.7   & 57.9 &  60.3    \\
    {AAA}                       & 61.2  & 54.1 &   57.7 \\
    {multi-channel+AAA}         & 62.5   & 57.6 &  60.1    \\
    {F2A+AAA}                   & \textbf{64.4}  & \textbf{60.3} &   \textbf{62.4}  \\
    {F2A+AAA+local-AU}          & 63.3  & 59.7 &   61.5  \\\midrule
    {F2A+AAA+$l_{ma}$}          & 64.4  & 60.5 &    62.5 \\
    {F2A+AAA+$l_{div}$}         & 64.7  & 62.3 &   63.5\\
    {F2A+AAA+both}              & \textbf{64.9}   & \textbf{62.4} & \textbf{63.7}\\\bottomrule
\end{tabular}
}
\label{tab:ablation_study}
\end{center}
\end{table}

\begin{figure}
\begin{center}
\includegraphics[width=0.95\linewidth]{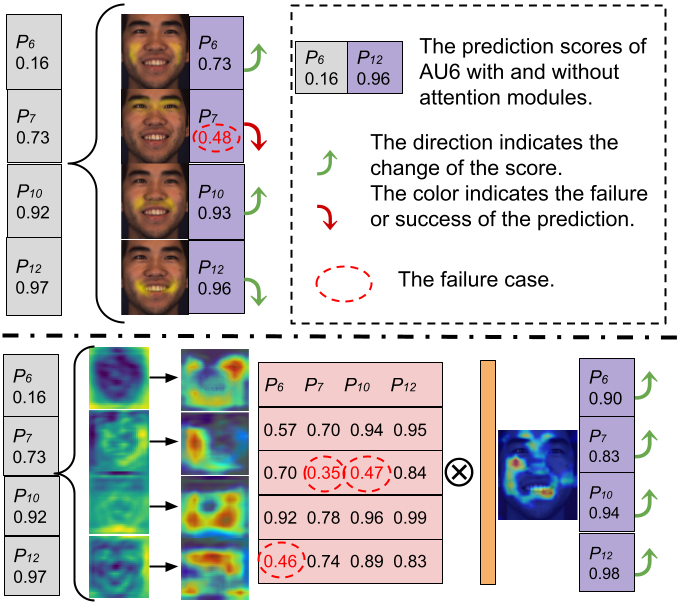}
\caption{Prediction score and visualization using different methods. The upper part is the conventional multi-attention for AU detection. The lower part is our method. Here it only shows the prediction confidence of AU6, AU7, AU10, and AU12 with positive ground truth. The number of attention channels is 7 but we only show 4 of them. By re-calibrating the prediction scores from multi-channels, our method effectively suppresses the failure cases, which is difficult for previous multi-attention models.}
\label{attention_map_2}
\end{center}
\end{figure}

\subsection{Visualization}

Fig.\ref{attention_map_2} illustrates how our method improves the prediction ability of attention learning in AU detection. The conventional multi-attention method, in Fig.\ref{attention_map_2} (up), disentangle a face into several AU-related branches and use partial attention to learn discriminative features. However, the prediction of each AU highly relies on the one-dimensional confidence table due to one-to-one mapping from attention to specific AU. This figure shows it achieves high confidence for AU6, AU10, and AU12 but gets a failure prediction for AU7. This is caused by the uncertainty of the attention learning mechanism. There is no additional measures to restore or cure a single attention branch with low quality. Fig.\ref{attention_map_2} (down) shows our method. We expand the one-dimensional prediction confidence table to a two dimensional one (the pink matrix). With the diverse representations of attention maps, our model can selectively choose which channel is more convincing and which one is not. Thus, the failure cases in the pink matrix has been effectively suppressed.

\begin{figure}
\begin{center}
\includegraphics[width=0.95\linewidth]{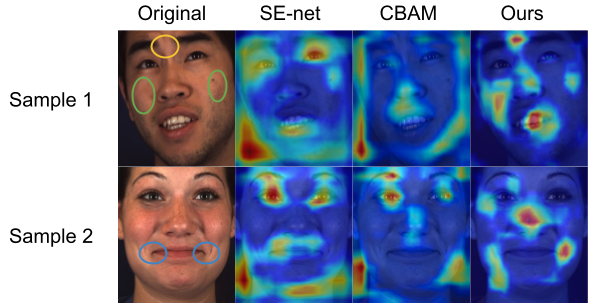}
\caption{Visualization of attention maps on BP4D. Each column shows the results using different models. The first row shows the original images. Yellow: AU4; Green: AU6; Blue: AU12, AU14, AU15}
\label{attention_map}
\end{center}
\end{figure}

To further elaborate on how the proposed approach affects the recognition task, we provide a qualitative analysis. We apply the Grad-GAM \cite{Experiments_29} to visualize the attention maps of two subjects in BP4D. We compare it with several attention models on BP4D for AU detection in Fig.\ref{attention_map}. Through observation, we found our method can concentrate more on AU-related regions. For AU4 ( Brow Lowerer) and AU6 (Cheek Raiser) in sample 1 and AU12 (Lip Corner Puller), AU14 (Dimpler), and AU15 (Lip Corner Depressor) in sample 2, our model performs better in focusing on the target area accurately than the channel-wise attention model (SE-Net) and Mixture attention model (CBAM). Besides, unlike the baseline attention models, our model alleviates the irrelevant regions(e,g. background area and pupil area) of the face in column two and column three. We argue that irrelevant areas can cause negative interference to facial parts localization and image detection.

\section{CONCLUSIONS AND FUTURE WORKS}

In this article, we explained the robustness issue of attention-based learning which is overlooked in facial action analysis before. In addition, we explored the dependency between deep feature maps and attention maps. We proved that the performance of neural attention, as a key factor of CNN, is influenced by the representational quality. Based on this, we proposed a novel self-diversified multi-channel attention to seek a more robust attention design on facial action unit detection and facial expression recognition. In order to sense and regulate the uncertainties of attention, we propose several key components that enhance the representational power of single facial attention, expand the diverse pattern of attention maps in receptive fields, and capture the ``Inter-attention" correlation for getting refined attention maps. The method is evaluated on six widely used benchmark datasets for both AU detection and facial expression recognition. It achieves superior performance over the peer state-of-the-art methods.

\section{ACKNOWLEDGMENTS}

The material is based on the work supported in part by the NSF under grant CNS-1629898 and the Center of Imaging, Acoustics, and Perception Science (CIAPS) of the Research Foundation of Binghamton University.


{\small
\bibliographystyle{ieee}
\bibliography{egbib}
}

\end{document}